\documentclass[letterpaper, 10 pt, journal, twoside]{IEEEtran}

\usepackage{graphicx} 
\usepackage{amsmath} 
\usepackage{amssymb}  
\usepackage{algorithm}
\usepackage{algpseudocode}
\usepackage{mathtools}
\usepackage{hyperref}

\ifCLASSOPTIONcompsoc
 \usepackage[caption=false,font=normalsize,labelfont=sf,textfont=sf]{subfig}
\else
 \usepackage[caption=false,font=footnotesize]{subfig}
\fi

\begin{document}

\title{Improved Learning of Robot Manipulation Tasks via Tactile Intrinsic Motivation}
\author{Nikola Vulin$^{1}$, Sammy Christen$^{1}$, Stefan Stev\v{s}i\'{c}$^{1}$ and Otmar Hilliges$^{1}$%
\thanks{$^{1}$AIT Lab, Department of Computer Science, ETH Zurich, Switzerland {\tt\footnotesize nikola.vulin@inf.ethz.ch}}%
\thanks{\copyright 2021 IEEE. Personal use of this material is permitted.  Permission from IEEE must be obtained for all other uses, in any current or future media, including reprinting/republishing this material for advertising or promotional purposes, creating new collective works, for resale or redistribution to servers or lists, or reuse of any copyrighted component of this work in other works.}
\thanks{This is the author’s version of an article that has been published in IEEE Robotics and Automation Letters. The final version of record is available at \url{https://dx.doi.org/10.1109/LRA.2021.3061308}.}
}




\maketitle

\begin{abstract}
In this paper we address the challenge of exploration in deep reinforcement learning for robotic manipulation tasks. In sparse goal settings, an agent does not receive any positive feedback until randomly achieving the goal, which becomes infeasible for longer control sequences. Inspired by touch-based exploration observed in children, we formulate an intrinsic reward based on the sum of forces between a robot's force sensors and manipulation objects that encourages physical interaction. Furthermore, we introduce contact-prioritized experience replay, a sampling scheme that prioritizes contact rich episodes and transitions. We show that our solution accelerates the exploration and outperforms state-of-the-art methods on three fundamental robot manipulation benchmarks.
\end{abstract}


\IEEEpeerreviewmaketitle

\section{INTRODUCTION}
\IEEEPARstart{M}{odel-free} deep reinforcement learning (DRL) algorithms have demonstrated great potential in solving sequential decision making problems, such as learning to play video games in Atari \cite{mnih2015human}, defeating the world champion in the game of Go \cite{silver2017mastering}, or controlling robotic systems for locomotion \cite{hwangbo}, navigation \cite{christen2021} and manipulation \cite{akkaya2019solving}. For complex tasks and real-world scenarios, in particular in robotics, formulating a feasible reward function is inherently difficult and tedious. Hence, reward functions are most easily formulated as a sparse signal received upon reaching the final goal. This leads to inefficient exploration, because an agent receives crucial feedback rarely and in a delayed manner. Reward functions are typically based on reaching \emph{extrinsic} states. Thus, the reward does not leverage internal robot states or other internal representations. To address inefficient exploration, an \emph{intrinsic} reward inspired by intrinsic motivation observed in humans \cite{ryan2000intrinsic} can be added to the reward function. 

Humans strongly rely on the sense of touch for exploration and use it as guidance when interacting with their surroundings. For example, humans can manipulate objects even without visual feedback, and removing tactile cues significantly reduces manipulation capabilities \cite{jones2006human}. Furthermore, it has been shown that force feedback is used by infants to explore the physical properties of objects \cite{kosoy2020exploring}. Our insight is that force feedback can be leveraged in a general manner for learning a diverse set of manipulation tasks. Inspired by the touch-based exploration observed in small children, we propose an \emph{intrinsic} reward based on tactile signals to enhance exploration in robotic manipulation environments. Furthermore, we introduce a sampling scheme that encourages robot-object interaction via prioritization of meaningful trajectories.

Intrinsic rewards have been introduced for discrete domains based on state novelty \cite{bellemare2016unifying}, uncertainty \cite{pathak2017curiosity, burda2018exploration} or information gain \cite{baranes2013active}. Using methods based on intrinsic motivation accelerates the training of an agent, but requires an additional model to extract a reward from state information. In DRL approaches, tactile robot-object interaction information is often ignored, although this information might be valuable during exploration. Some prior works extend the state space with tactile information to provide explicit interaction cues. This results in improvements in terms of sample efficiency and robustness for grasping \cite{merzic2019leveraging}, in-hand manipulation \cite{melniktactile}, and human-robot interaction \cite{christen2019}. However, these methods modify only the state space \cite{merzic2019leveraging, melniktactile}, or focus on a set of specific tasks \cite{merzic2019leveraging, christen2019}.

\begin{figure}[t]
    \centering
    \includegraphics[width=0.9\columnwidth]{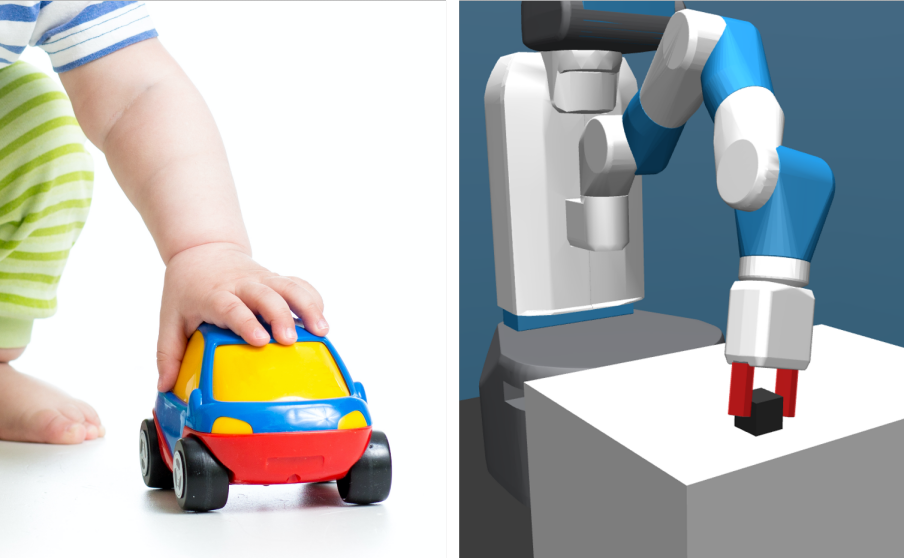}
    \caption{Inspiration of our work comes from intrinsic motivation of infants exploring the world through physical interactions. We model tactile sensing with force sensors (red) and leverage tactile information to accelerate learning.}
    \label{fig:teaser}
\end{figure}

In our method (see Figure \ref{fig:teaser}), we add force sensors to the end-effector and extend the state space with force measurements, similarly to other work \cite{merzic2019leveraging, melniktactile, christen2019}. We then go beyond prior work and introduce an intrinsic reward to guide the exploration. Our intrinsic reward behaves like an intermediate, simpler to achieve, directive base. Having an intermediate base can improve exploration, since the agent returns to this good position and does not lose track due to probabilistic exploration \cite{ecoffet2019go}. The reward motivates the agent to touch the object while allowing it to explore different interactions, i.e., when and how long to touch an object.  We experimentally show that our reward leads to significantly better performance on benchmark manipulation tasks, especially when the task difficulty is increased. 

Another way of addressing exploration is to use data augmentation techniques such as Hindsight Experience Replay (HER) \cite{andrychowicz2017hindsight}, which replaces a task-specific goal with a state that was achieved in hindsight of a sampled transition. This will induce intermediate rewards for failed trials. However, HER randomly samples states from the replay buffer, including trajectories where the manipulation object does not move, which may slow down learning.
We therefore introduce a replay buffer sampling scheme that allows an agent to prioritize meaningful trajectories. More specifically, we first modify the sampling probability of entire episodes depending on whether contacts have happened. From a batch of selected episodes, we sample virtual goals, with an increased weight on transitions that occur after initial contact. We then go back in time to find training transitions, which narrows down the search space to meaningful transitions for manipulation tasks. We show empirically that this leads to faster convergence compared to HER.

The main contributions of this paper can be summarized as follows: i) An \emph{intrinsic} force-based reward that accelerates the exploration and improves the performance on several benchmark robotic manipulation tasks. ii) A novel sampling scheme for off-policy algorithms in contact-based tasks to increase the sample efficiency. iii) We provide a detailed evaluation of our approach and comparison with the baselines. We analyze the contributions of the method components in an ablation study and show that our method performs significantly better, especially with increased task difficulty. 

\section{RELATED WORK}

\subsection{Tactile Feedback}

For humans, tactile sensing is an important sensory modality, besides visual perception, when manipulating objects. It is rich in information and free of external disturbances, compared to visual sensing that is influenced by occlusions and poor lighting conditions. While the majority of DRL works do not consider tactile feedback for learning robotic manipulation tasks, a few have shown the benefits and potential of considering this additional sensory modality \cite{merzic2019leveraging, melniktactile, christen2019, huang2019learning, wu2020mat}. 

In \cite{merzic2019leveraging}, the authors show that tactile sensing can improve the robustness of grasping under noisy position sensing, thus supporting exact position estimation. \cite{melniktactile} demonstrates that tactile sensing provides the agent with essential information and makes the learning more sample efficient for in-hand manipulation tasks. Unlike our method, these approaches only add the force measurement to the state space.

More similar to our method, \cite{christen2019} and \cite{huang2019learning} extend their state space with tactile sensing and use it in the reward function. To learn human-robot interactions, \cite{christen2019} propose a specific reward to ensure contact between the human and the robot. \cite{huang2019learning} uses the feedback of tactile sensors as a penalty to avoid high impacts and hence to learn gentle manipulation. However, in both methods the reward is used to trigger a task-specific behavior. In contrast, we formulate an intrinsic reward to encourage general task-independent exploration.

\subsection{Intrinsic Motivation}

Humans are efficient learners, showing intrinsically motivated behaviors when exploring the world around them \cite{ryan2000intrinsic}. Inspired by human exploration strategies, intrinsic motivation in reinforcement learning (RL) has been used to improve exploration in sparse reward settings \cite{aubret2019survey}. The research mostly relies on intrinsic rewards that an agent collects for visiting novel states. This type of reward can be implemented by computing pseudo-counts through density models \cite{bellemare2016unifying} or by estimating state novelty with neural networks \cite{burda2018exploration}. Furthermore, one can model curiosity to guide the agent towards areas where the uncertainty of the agent's prediction of the following states is high \cite{pathak2017curiosity}. These methods need to estimate additional, computationally expensive values, which slows down training. In contrast, our intrinsic reward only relies on tactile feedback present in the state space.

\subsection{Experience Replay}
\label{sec:experiencereplay}

Experience replay \cite{lin1992self} is applied in off-policy DRL algorithms to store experience transitions in a buffer and reuse them during training. Popularized in DQN \cite{mnih2015human}, it has become a standard procedure to mitigate sample efficiency and stabilize training in most state-of-the-art off-policy algorithms \cite{lillicrap2015continuous}. An extension includes Prioritized Experience Replay (PER) \cite{schaul2015prioritized}, where transitions are prioritized based on how valuable they are to the learning process via the temporal difference error. 

For the case of learning from sparse rewards in continuous control problems, Hindsight Experience Replay (HER) \cite{andrychowicz2017hindsight} has proposed a data augmentation technique that achieves large performance improvements over standard approaches. The transitions stored in the buffer are used for training, with the difference that rewards are computed with respect to virtual goals. Even when the agent fails to complete the task, virtual goals provide a way to achieve intermediate rewards. HER was extended in \cite{zhao2018energy} by prioritizing episodes based on the sum of kinetic and potential energy of the manipulation object. This can lead to prioritization of undesired trajectories, because the faster the object moves, the higher the priority, which may not be what is beneficial for the task at hand. In contrast, our prioritization scheme treats episodes with contact more equally and also modifies the sampling probability of training transitions within an episode.

\section{PRELIMINARIES}

\newcommand{\StatesSet}{\mathcal{S}}
\newcommand{\ActionsSet}{\mathcal{A}}
\newcommand{\TransitionSet}{\mathcal{T}}
\newcommand{\GoalsSet}{\mathcal{G}}
\newcommand{\state}[1]{{s}_{#1}}
\newcommand{\goal}[1]{{g}_{#1}}
\newcommand{\action}[1]{{a}_{#1}}
\newcommand{\reward}{r}
\newcommand{\discountRate}{\gamma}
\newcommand{\QFunction}{Q}
\newcommand{\policy}{\pi}
\newcommand{\Cost}{J}
\newcommand{\mdp}{\mathcal{M}}

\subsection{Problem Definition}

In this paper, we focus on using reinforcement learning to solve robotic manipulation tasks, where a robot needs to interact with an object to complete the task. For example, the robot may need to pick objects and place them in a container, build a structure, or use tools. A straightforward way to learn these tasks is to reward the agent once the task is completed. In contrast, shaping the reward might be very difficult because there are multiple intermediate steps that the robot needs to achieve before it can reach the final goal. When sparse rewards are used, no signal directs the agent towards the goal, and hence the agent reaches it only by chance via random exploration. Thus, algorithm convergence is usually very slow, or the algorithm does not even converge if the task is too complex. We aim to improve exploration in robotic manipulation tasks and hence learn the required skills faster. In manipulation tasks, we can often resort to the goal-conditioned formulation of the reinforcement learning problem. This can help to mitigate the exploration problem by applying methods such as HER \cite{andrychowicz2017hindsight}.

\subsection{Goal-conditioned Deep Reinforcement Learning}
\label{sec:drl}

To model our control problem, we use the standard formulation of a Markov Decision Process (MDP) and extend it with a set of goals $\GoalsSet$. An MDP is defined by a tuple $\mdp = \{\StatesSet, \ActionsSet, \GoalsSet, \mathcal{R}, \TransitionSet, \rho_0, \discountRate \}$, where $\StatesSet$ is the state set, $\ActionsSet$ the action set,  $\mathcal{R}_t=r(s_t,a_t, g_t)$ the reward function,  $\TransitionSet = p(\state{t+1}|\state{t}, \action{t})$ the transition dynamics of the environment  with $\state{t} \in \StatesSet$ and $\action{t} \in \ActionsSet$,  $\rho_0 = p(\state{1})$ the initial state distribution, and $\gamma$ a discount factor. Our aim is to maximize the expected sum of discounted future rewards with a policy $\pi: \StatesSet \times \GoalsSet \rightarrow \ActionsSet$, which is a mapping from states and goals to actions. We adopt the actor-critic framework and use the Q function to describe the expected future reward:  

\begin{equation}
    Q(s_t,g_t, a_t) = \mathbb{E}_{a_t \sim \pi, s_{t+1} \sim \mathcal{T}} \left [\sum_{i=t}^{T}\discountRate^{i-t}\mathcal{R}_t \right ]
    \label{eq:q_func}
\end{equation}

We approximate both the value function (critic) and the policy (actor) with neural networks and use the DDPG algorithm \cite{lillicrap2015continuous} for training. Thus, the critic network is parameterized by $\theta^Q$ and minimizes the following loss:

\begin{equation}
\begin{split}
&L(\theta^Q) = \mathbb{E}_{\mdp} \left [ \left( Q(\state{t}, \goal{t}, \action{t} ; \theta^Q) - y_t \right)^2   \right],\\
&y_t = \reward(\state{t}, \goal{t}, \action{t}) + \discountRate Q(\state{t+1}, \goal{t+1}, \action{t+1} ; \theta^Q).
\label{eq:q_bellman}
\end{split}
\end{equation} 

The policy network uses parameters $\theta^{\policy}$ and is trained to maximize the Q-value:
\begin{equation}
\medmuskip=0mu
\thinmuskip=0mu
L(\theta^{\policy}) = \mathbb{E}_{\pi} \left [Q(\state{t}, \goal{t}, \action{t} ; \theta^Q)    \vert \state{t}, \goal{t}, \action{t} = \policy(\state{t}, \goal{t}; \theta^{\policy}) \right ]
\label{eq:pi}
\end{equation}

The algorithm is off-policy and therefore uses an experience replay buffer for training. The data is collected in episode rollouts (we use the word trajectory interchangeably in this paper) and stored as transitions $(s_t, g_t, a_t, r_t, s_{t+1})$ in the buffer. These transitions are then sampled from the replay buffer to train the neural networks that approximate the Q function and control policy $\pi$.

\subsection{Hindsight Experience Replay (HER)}
\label{sec:her}
Hindsight Experience Replay (HER) \cite{andrychowicz2017hindsight} is a data augmentation technique that significantly improves the exploration of goal-conditioned reinforcement learning algorithms. In HER, the key idea is to learn from failed trials. Therefore, uniformly sampled transitions from the replay buffer are modified by replacing the goal $g_t$ of a transition with a state achieved in a subsequent time step of the same trajectory. Accordingly, the reward $r_t$ must be recalculated, which will consequently induce rewards for trajectories that failed to reach the original goal. From these unsuccessful trials, the goal-conditioned policy should learn to reach the original goal more efficiently via extrapolation. However, the goals used in HER are usually object-centric. Hence, it can take a significant amount of trials until the agent starts to move the object and virtual goals start to affect the learning algorithm. This is a substantial downside of HER, which we address with our method.

\subsection{Simulation Environments}
\label{sec:simulation}

We evaluate our method on three different benchmark tasks from the robotics collection \cite{plappert2018robotics} of OpenAI Gym \cite{brockman2016openai}. The environments are based on the physics engine MuJoCo \cite{todorov2012mujoco}. MuJoCo allows to design tailored physical environments and computes the dynamics and contact interactions between rigid bodies. We use a 7 DoF robot manipulator (see Figure \ref{fig:teaser}). Marked in red are the locations of the force sensors, which are placed on the end-effector. The robot is controlled via torque actions that are sent to its motor joints.

\begin{figure}[t]
    \centering
    \includegraphics[width=0.98\columnwidth]{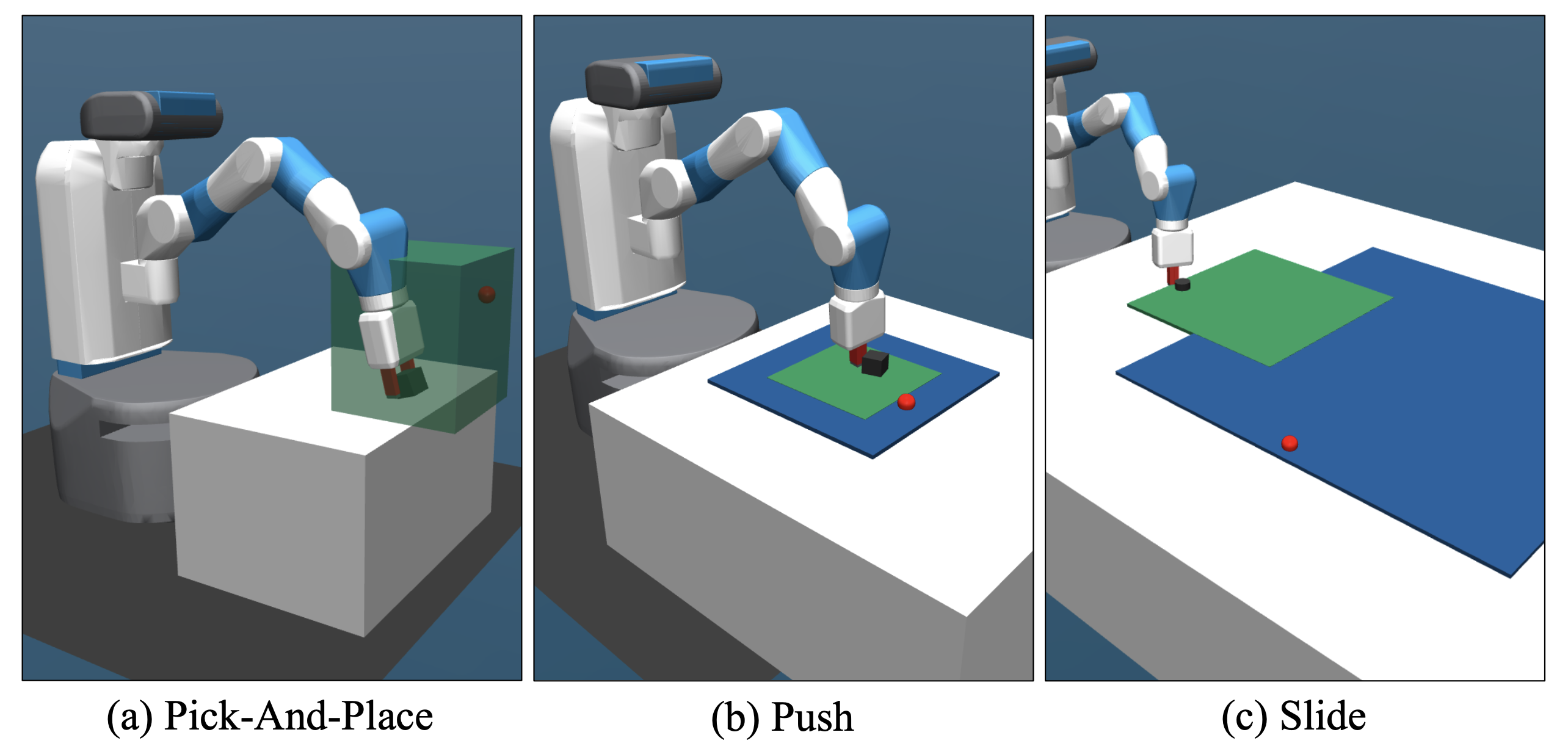}
    \vspace{-0.2cm}
    \caption{Overview of the benchmark tasks. We enlarge the benchmark's {\cite{plappert2018robotics}} original goal space (green) to a larger area (blue) to increase the task difficulty. For the \textit{Pick-And-Place} task, we sample goals uniformly contrary to the original benchmark task, which uses 50\% of targets on the table.}
    \label{fig:simulation}
\end{figure}

Figure \ref{fig:simulation} shows the three tasks considered: \textit{Pick-And-Place}, \textit{Push} and \textit{Slide}. These tasks represent a diverse set of manipulation tasks. In \textit{Pick-And-Place}, the goal is to pick up an object and move it to a target. In \textit{Push}, the task consists of pushing an object on a surface to a target position. Contrarily, in \textit{Slide} the robot has to kick a puck in order to reach a distant goal that is out of reach of the robot's workspace. The red sphere in Figure \ref{fig:simulation} marks the target position an object needs to reach. The original goal space \cite{plappert2018robotics} is illustrated in green. To increase the difficulty of the problem, we enlarge the spread of the goal space from the green to the blue area. For \textit{Pick-And-Place}, the original problem \cite{plappert2018robotics} uses the simplification that 50 \% of goals are on the table and not in the air. This can lead to instances where the robot does not need to learn to grasp the object in order to complete the task. Hence, we modify the task such that all goals are sampled uniformly in the marked green area. 

\section{METHOD}

\begin{figure*}
    \centering
    \includegraphics[width=0.7\textwidth]{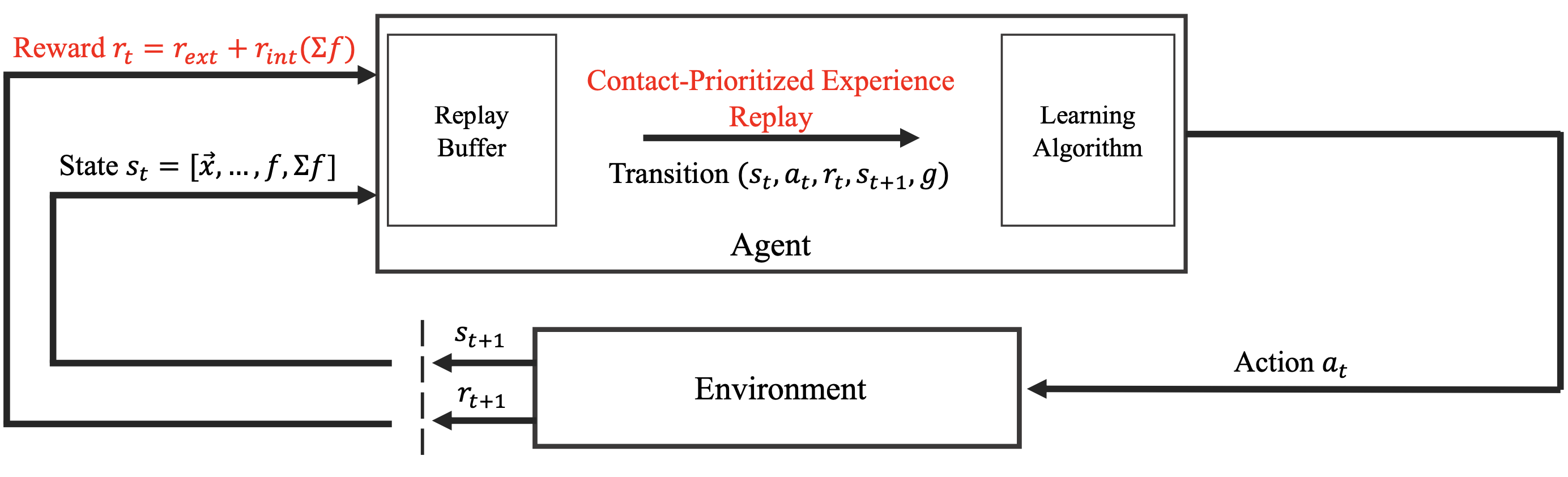}
    \vspace{-0.2cm}
    \caption{The standard reinforcement learning loop and our two method extensions in red. The reward function is extended by an intrinsic reward and a sampling scheme that prioritizes contact-rich transitions is introduced.}
    \label{fig:method_overview}
\end{figure*}

When training robots to execute manipulation tasks via reinforcement learning, the reward function is most often defined as sparse. Therefore, it is essential to have an efficient exploration approach. We propose a method that improves exploration efficiency by leveraging force sensor information in the goal-conditioned reinforcement learning setting (see Figure~\ref{fig:method_overview}).

Our main insight is that sense of touch can guide robots to take actions that result in manipulation of an object. To implement the sense of touch, we extend our state-space with force measurements from tactile sensors positioned at the end-effector \cite{merzic2019leveraging, melniktactile, christen2019}. We introduce a touch-based intrinsic reward function to direct exploration towards the states where the robot touches the object. The trajectory in the state space that leads to the task solution passes through the state where the robot touches the object. Hence, incentivizing the agent to reach such states will guide the exploration towards the final task goal. Using the intrinsic reward in combination with HER utilizes virtual goals much earlier because the robot starts to move the object significantly faster. We further exploit the touch information by introducing a Contact-Prioritized Replay Buffer (CPER). CPER uses contact clues to sample more informative transitions and virtual goals for training the Q function and control policy $\pi$, which further speeds up the learning.

\subsection{State Space}
\label{sec:state_space}

We extend our state space with force measurements from tactile sensors positioned at the end-effector \cite{merzic2019leveraging, melniktactile, christen2019}. The original state space consists of a mixture between proprioceptive state information, i.e., joint positions and joint velocities, and relevant information about the manipulation object, such as the object's velocity and the position $x_{\text{obj}}$, which we will use in the reward function. We further extend the state space with the measured force $f_t$ at the end-effector and the sum of all measured forces until the current step $\sum^t_{i = 0} f_i$. The accumulated force is a good measure to track how much contact occurred during an episode, which is what our intrinsic reward is based on.

\newenvironment{test}{%
\setlength{\dbltextfloatsep}{0pt}
}

\begin{algorithm*}[tb]
\caption{Contact-Prioritized Experience Replay (CPER)}\label{alg:ours}
\begin{algorithmic}[1]
\State \textbf{Given:} Transitions $(s_t, g_t, a_t, r_t, s_{t+1})$ of episode $e \in E$ and step $t \in T$ stored in Replay Buffer $\mathcal{R}$

\State Compute probability distribution $p_{\text{transition}}(e, t)$ \Comment{Fig. \ref{fig:method}, Eq. \ref{eq:ptrans}}

\State Sample mini-batch $\mathcal{B}$ of episodes from $\mathcal{R}$ with marginal probability distribution $p_{\text{episode}} = \sum_{t=0}^T p_{\text{transition}}(e, t)$ 

\For {episode $e_b$ in $\mathcal{B}$}
    \State Sample state $s_{t'}$ from $p_{\text{transition}}(e_b, t)$
    \State Sample transition $(s_t, g_t, a_t, r_t, s_{t+1})$ back in past where $0 \leq t \leq t' $.
    \If{\text{Hindsight Transition}} \Comment{See HER \cite{andrychowicz2017hindsight}}
        \State Replace original goal with sampled state $g_t \gets s_{t'}$
        \State Recompute $r' := r(s_t, a_t, g_{t})$
    \EndIf
\EndFor

\State Perform one step of optimization using Algorithm $\mathbb{A}$ and mini-batch $\mathcal{B}$
\end{algorithmic}
\end{algorithm*}

\subsection{Intrinsic Force Reward (IR)}
\label{sec:forcereward}

We extend an extrinsic reward function $r_{\text{ext}}(s, g)$ with our intrinsic force reward $r_{int}(s)$, which is independent of the goal:
\begin{equation}
r(s, g) = \omega_{\text{ext}} * r_{\text{ext}}(s, g) + \omega_{\text{int}} * r_{\text{int}}(s),
\end{equation}
where $\omega_{\text{ext}}$ and  $\omega_{\text{int}}$ are the weights on the extrinsic and intrinsic reward function, respectively.
We empirically found that the weights $\omega_{\text{ext}} = 0.75$ and $\omega_{\text{int}} = 0.25$ provide a good balance between our intrinsic reward as an auxiliary reward, while putting more emphasis on the extrinsic reward function. 
The extrinsic reward function is formulated sparsely, where the indicator function $\mathbb{I}$ returns a positive signal of magnitude 1 if the object's position $x_{\text{obj}}$ is within the range $\varepsilon_{\text{pos}}$ around the goal $g$: 
\begin{equation}
r_{\text{ext}}(s, g) = \mathbb{I}\bigg[ \big\| g- x_{\text{obj}} \big\| < \varepsilon_{\text{pos}} \bigg]
\label{eq:ext_reward}
\end{equation}
In all other cases, the reward is 0. The intrinsic reward depends only on the state and returns a reward if the current sum of forces $\sum^t_{i = 0} f_i$ is higher than a minimal amount of desired contact interaction $\varepsilon_{\text{force}}$: 
\begin{equation}
r_{\text{int}}(s_t) = \mathbb{I}\bigg[ \sum^t_{i = 0} f_i  > \varepsilon_{\text{force}} \bigg] 
\label{eq:intr_reward}
\end{equation}
The intrinsic force reward allows the agent a certain freedom about when and how long to touch the object. As soon as a minimal amount of contact occurred, i.e., when $\Sigma f$ is greater than a minimal threshold, the agent receives the intrinsic reward until the end of the episode and can concentrate on solving the extrinsic task. The reward encourages the agent to manipulate the object and to bring it into motion. Therefore, the value of the threshold $\varepsilon_{\text{force}}$ should be chosen large enough to avoid falsely providing the intrinsic reward due to sensor noise. We empirically set the threshold to $10N$ for the tasks \emph{Push} and \emph{Pick-And-Place} and $3N$ for the \emph{Slide} task, since this environment requires only a short interaction with the object. However, we found that the threshold is robust across a range of different values.

We formulate the intrinsic reward such that it has a minimal influence on the final task. After the force threshold is reached, the agent can freely explore interaction with an object. Hence, we  direct the agent to a crucial intermediate base from where the goal can be reached more quickly. Importantly, we use the same reward function across multiple tasks, demonstrating that our formulation can work in the general case. 
We also address the issues of HER in early exploration. Specifically, when an object is not moved during an episode, HER does not have an effect on training (see Section \ref{sec:her} for details). Our intrinsic reward circumvents the early exploration issue by guiding the agent to interact with the object and significantly speeds up convergence as we show in Section \ref{sec:exp_robot_manipulation}.

\subsection{Contact-Prioritized Experience Replay (CPER)}
\label{sec:sampling}

When training the value and policy networks, transitions are sampled from a replay buffer. The standard procedure is to sample the transitions uniformly. However, not all transitions are of the same value to the learning algorithm. Therefore, prioritizing more informative transitions helps to speed up learning \cite{schaul2015prioritized}. Furthermore, the selection of virtual goals is important as well. We leverage tactile information to alter the sampling of transitions and virtual goals. 

We therefore introduce Contact-Prioritized Experience Replay (CPER), which is a sampling prioritization scheme under the assumption that contact rich trajectories include more meaningful information for the learning process. Our method relies on two important features to accelerate learning. First, we prioritize trajectories where the agent touches the object. These trajectories contain important information about the relevant motions of the agent and the object. Other trajectories include robot motions where the end-effector may be far away from the object and are therefore less relevant for training. Second, we sample for virtual goals first and then find training transitions by sampling from previous time steps. 
We sample virtual goals from states occurring after the contact to induce meaningful intermediate rewards. By sampling transitions from previous time steps, we increase the chance of finding relevant transitions. This reversed scheme helps in tasks like \emph{Slide}, where most important transitions occur prior to the contact phase. We experimentally show that using CPER leads to significantly faster training compared to the case when only our intrinsic reward is used. We now explain our sampling scheme in more detail.

\begin{figure}[t]
    \centering
    \includegraphics[width=0.95\columnwidth]{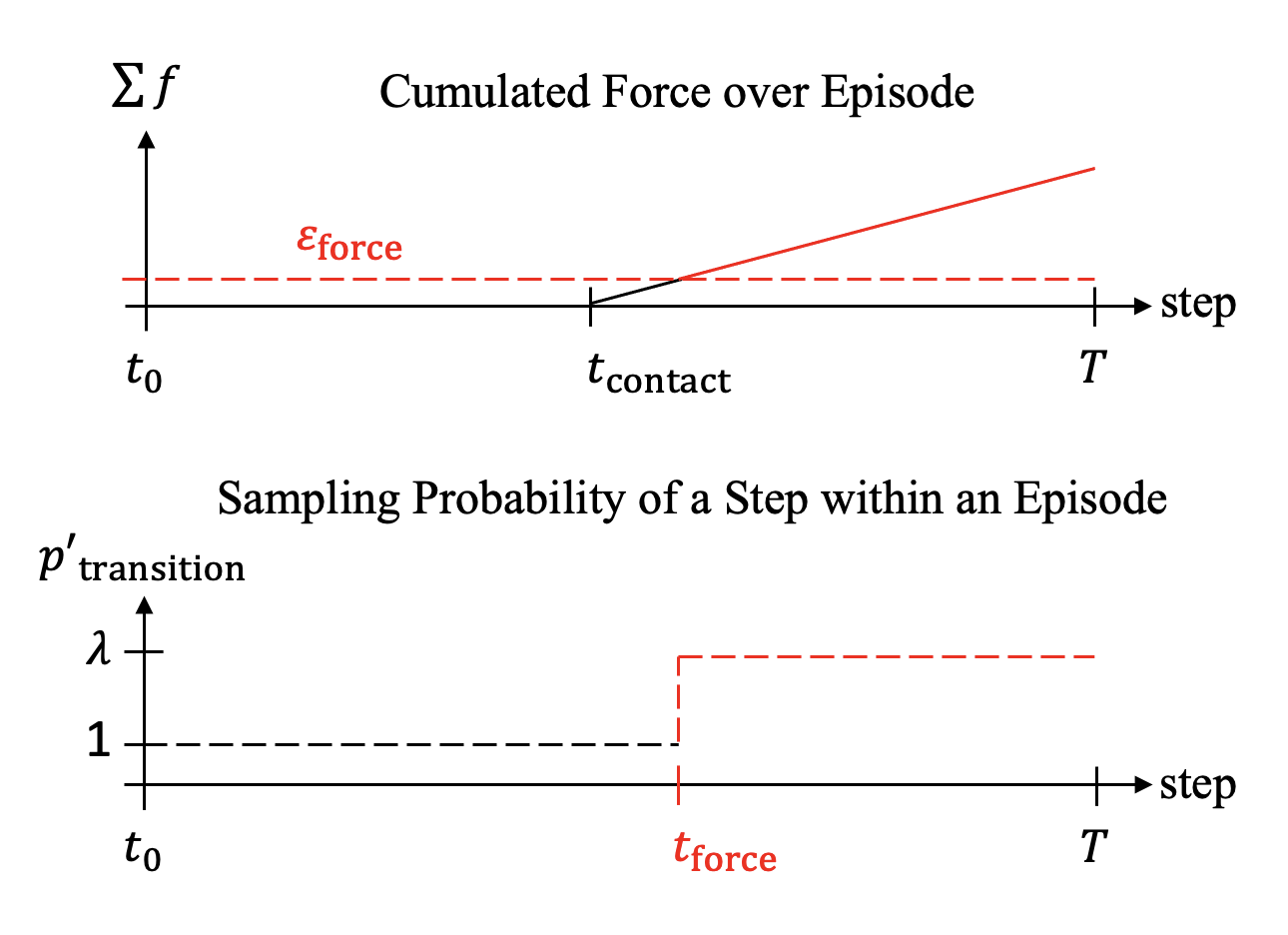}
    \vspace{-0.5cm}
    \caption{Computation of $p'_{\text{transition}}$. Once the force threshold $t_{\text{force}}$ is surpassed, the probability is multiplied by a factor $\lambda$.}
    \label{fig:method}
    \vspace{-0.2cm}
\end{figure}

First, we compute the probability distribution $p_{\text{transition}}$ using $p'_{\text{transition}}$ (see Figure \ref{fig:method}) and the following equation:
\begin{align}
\begin{split}
p'_{\text{transition}}(e, t) &= \left\{\begin{array}{rl}
1 & \text{if } \sum\limits_{i=0}^t f_{i} <\varepsilon_{\text{force}} \\
\lambda & \text{otherwise},
\end{array}\right.\\[2ex]
p_{\text{transition}}(e, t) &= \frac{p'_{\text{transition}}(e, t)}{\sum_{e=0}^E\sum_{t=0}^T p'_{\text{transition}}(e, t)}
\label{eq:ptrans}
\end{split}
\end{align}

Once the sum of forces reaches the threshold $\varepsilon_{\text{force}}$ at time step $t_{\text{force}}$ within a trajectory, we increase the sampling probability of transitions for all subsequent time steps until the end of the episode $T$ by factor $\lambda$. We empirically found that a factor of 10 leads to the best performance.

Algorithm \ref{alg:ours} describes our method in detail and consists of two parts. First, we compute the marginal probability distribution $p_{\text{episode}}$ from the sum of $p_{\text{transition}}$ per episode in the buffer, which overweights trajectories where contact occurred and further prioritizes based on the time step of the first contact (lines 2-3). This induces a prioritization of episodes with contact-rich information. We then sample a batch of trajectories $\mathcal{B}$ according to $p_{\text{episode}}$ (line 3).

\begin{figure*}[t]
    \centering
    \includegraphics[width=0.98\textwidth]{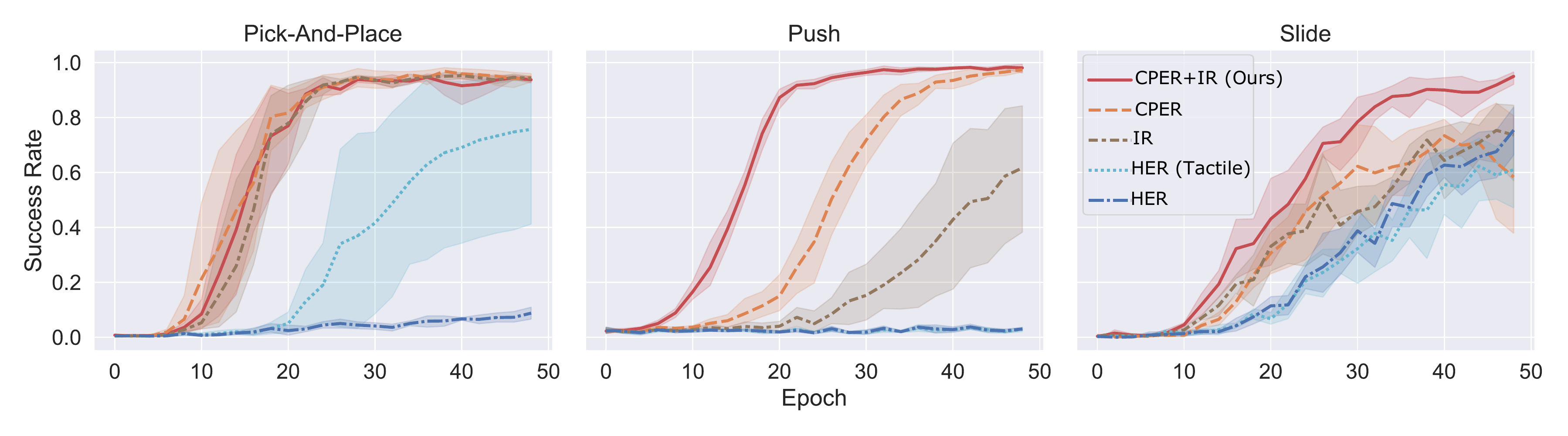}
    \vspace{-0.5cm}
    \caption{Learning curves for all three tasks. We show the average success rates over 5 random seeds with the corresponding confidence interval. 1 epoch corresponds to 40 training episodes. We show our full method with contact-prioritized experience replay and the intrinsic reward (CPER+IR) (red) and two ablations: CPER (orange) and IR (brown). We compare it with two baselines: HER with tactile sensing (light blue) and the original HER method (blue).}
    \label{fig:mainresult}
\end{figure*}

In the second part (lines 4-11), we sample a virtual transition at time $t'$ within an episode from batch $\mathcal{B}$ according to our modified distribution $p_{\text{transition}}$ (line 5). We then reverse back in time to find a training transition $(s_t, g_t, a_t, r_t, s_{t+1})$ and replace the original goal $g_t$ with the achieved state $s_{t'}$ from the virtual transition (line 6-9). Our sampling scheme allows us to find more meaningful virtual goals, narrowing down the search space to transitions that are relevant for robot manipulation tasks. In contrast to HER \cite{andrychowicz2017hindsight}, where first transitions are sampled uniformly and then virtual goals are selected from achieved states in the near future, we search for relevant goals first and then sample back in time to find training transitions. Since we first sample goal transitions with a higher sampling probability for transitions after the first contact (see Equation {\ref{eq:ptrans}}), it ensures that states in the trajectory before the contact are not neglected. We only replace the goals for a subset of transitions (80\%, the same as in {\cite{andrychowicz2017hindsight}}, line 7), keeping the original goal in the other subset as the task's final goal. 

\section{RESULTS}
\label{sec:experiments}

\subsection{Robot Manipulation Experiments}
\label{sec:exp_robot_manipulation}

To evaluate the proposed approach, we choose three standard manipulation tasks from MuJoCo benchmarks: \emph{Pick-And-Place}, \emph{Push}, and \emph{Slide} (see Section \ref{sec:simulation}). These three environments require significantly different interactions between the robot and the object, allowing us to evaluate the generalization abilities of our approach. The \emph{Pick-And-Place} task requires the robot to grasp the object, the \emph{Push} task requires one side touching, while \emph{Slide} requires a short kick interaction. In this experiment, we use the hardest version of each task as described in Sec.~\ref{sec:simulation}. Our goal is to evaluate our contributions in the most demanding settings. We use DDPG \cite{lillicrap2015continuous} with HER \cite{andrychowicz2017hindsight}, which we call only HER for abbreviation going forward, as a baseline because it is the state-of-the-art method for training goal-conditioned control policies. In the ablative study, we test how each of the components of our method performs. First, we extend the state space with force and cumulative force sensor readings, as explained in Sec.~\ref{sec:state_space}, and use HER to train the agent (HER~(Tactile)). Next, we separately add our intrinsic reward (IR) and our contact-prioritized experience replay (CPER), where we use our prioritization scheme with only the extrinsic reward. Finally, our full method uses all the components, combining the intrinsic reward and CPER (CPER+IR (Ours)).

Figure \ref{fig:mainresult} shows the results of the experiments. Our approach learns to complete the task significantly faster than HER. The difference is more prominent in \emph{Pick-And-Place} and \emph{Push}, since HER has problems solving these tasks due to the increased task difficulty. In the original settings \cite{plappert2018robotics}, HER manages to learn these tasks. However, our method leads to faster training in the original settings as well (see Figure~\ref{fig:workspace}). In all tasks, our approach converges after 20 to 30 epochs. In particular, at the beginning of training, our method finds successful actions faster. Hence, it yields a steeper learning curve, which indicates that our method motivates the agent to manipulate the object from the beginning.

Our ablation study shows that simply adding tactile information to the state space does not automatically lead to faster training. HER usually performs the same with and without the tactile sensor readings. In \emph{Pick-And-Place}, the tactile feedback improves the performance, but it is unstable. The faster convergence may imply that the tactile feedback gives an important signal that the object is grasped. When the intrinsic reward is used (IR), the performance is significantly improved compared to both HER and HER~(Tactile). This confirms that the proposed intrinsic reward function indeed speeds up training. In \emph{Slide}, the intrinsic reward has the smallest effect because this task demands a precise movement of the robot arm prior to the short kick of the object. 

Finally, we analyze the influence of our sampling scheme CPER. While there is no significant difference between IR and CPER+IR for \emph{Pick-And-Place}, we observe that it is crucial to choose transitions with valuable information in \emph{Push} and \emph{Slide} to achieve faster convergence. In \emph{Pick-And-Place}, the agent starts to grasp the object relatively fast due to the intrinsic reward, and hence most of the trajectories are prioritized for sampling. Therefore, our sampling scheme does not have a significant effect. Furthermore, applying CPER without IR shows similar results as the full method in \emph{Pick-and-Place}, implying that the intrinsic reward can potentially be avoided in some tasks. 

\subsection{Task Difficulty}
\label{sec:task_difficulty}

To study the effects of increasing the task difficulty, we conducted experiments using our method and the HER baseline. The original environments use a much smaller goal space than the robot can potentially reach in \emph{Push} and \emph{Slide}, or use the simplification that 50\% of the goals in \emph{Pick-And-Place} are on the table, which is a curriculum that simplifies the exploration. We test both approaches in three environments where we gradually increase the difficulty. In the Simple environment, we use the original settings. For Intermediate, we use settings that are half-way between the Simple and Hard environments, i.e., we increase the range of the goal space for \emph{Push} and \emph{Slide}, while we decrease the percentage of goals on the table for \emph{Pick-And-Place}. Hard is our final environment described in Sec.~\ref{sec:simulation}.

\begin{figure}
    \centering
    \includegraphics[width=0.98\columnwidth]{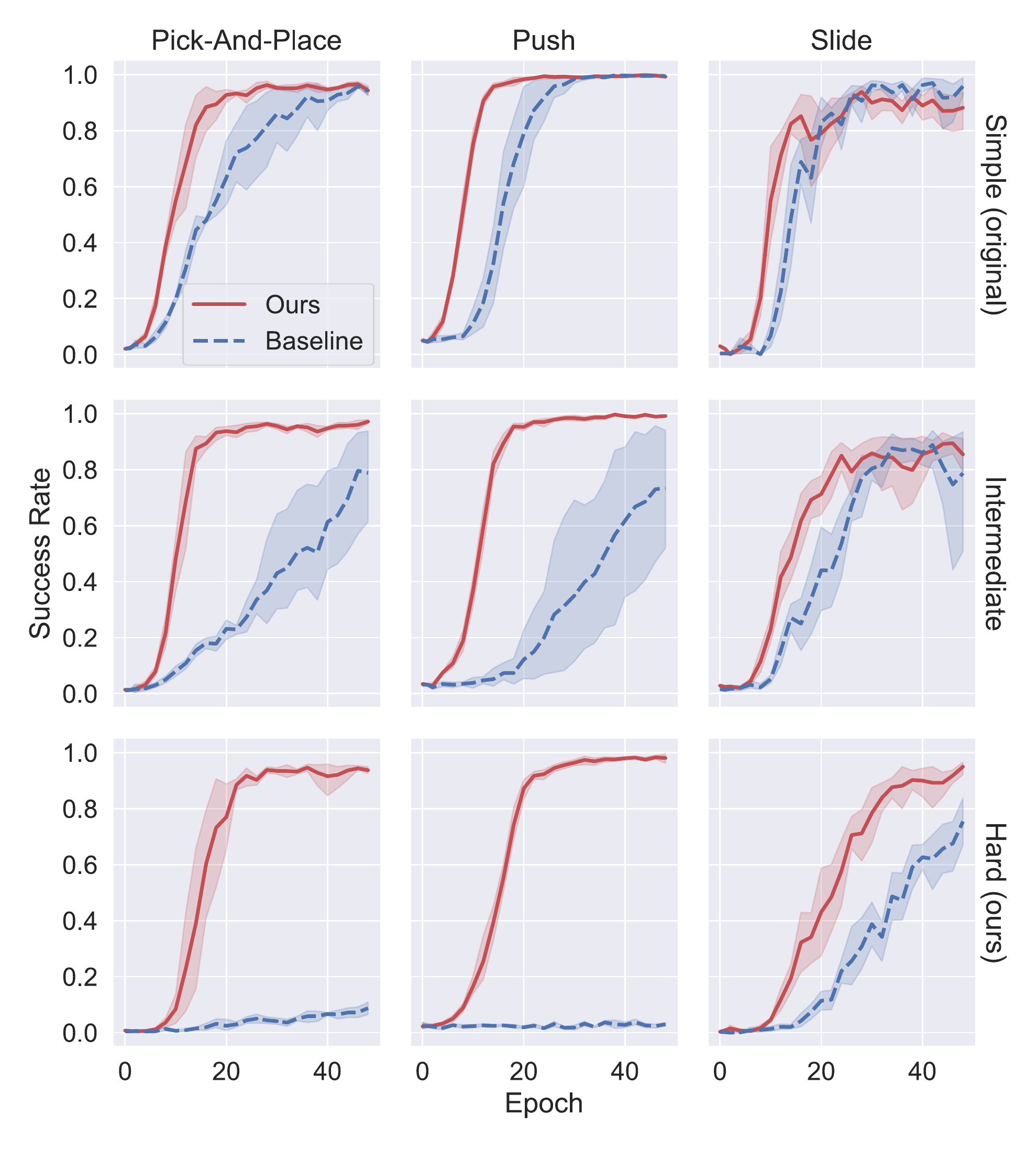}
        \vspace{-0.4cm}
    \caption{Learning curves for all three tasks and difficulty levels. We show the average success rate over 5 seeds with the corresponding confidence interval. 1 epoch corresponds to 40 training episodes. We compare our full method CPER + IR (red) and the original HER method (blue).}
    \label{fig:workspace}
\end{figure}

The results are illustrated in Figure \ref{fig:workspace}. When the task difficulty is kept the same as in \cite{plappert2018robotics} (Simple, top row), both methods reach similar success rates. However, we can see that our method converges faster. Similar results are observed when the task difficulty is slightly increased (Intermediate, middle row), although the convergence of the baseline slows even more compared to our method. Lastly, if the difficulty is increased as described in Section \ref{sec:simulation}, our method shows significantly better performance both in terms of convergence and success rate. These results are in accordance with insights from {\cite{lewisworkspace2019}}, which demonstrates that methods often perform well due to constrained goal spaces. Because of the enhanced exploration, our method manipulates the object once discovered and can therefore more efficiently generalize to a larger goal space.

\subsection{Sampling Ablation Study}
\label{sec:ablation}

In this experiment, we analyze our prioritization scheme CPER in more detail. CPER uses two main components: Increasing the probability of sampling transition from episodes with contact, and sampling transitions and virtual goals inside the trajectory based on the contact occurrence. We compare full CPER to a sampling scheme that uses just the first component, i.e., we prioritize episodes based on contact occurrences.  Hence, we sample the virtual goal state $s_{t'}$ uniformly, instead of sampling it from $p_{\text{transition}}$ (compare line 5 in Algorithm \ref{alg:ours}). We name this sampling scheme \emph{episode sampling}. We also compare against a more conventional baseline, where we prioritize episodes based on the achieved reward and sample the transitions uniformly, which we call \emph{reward prioritization}. In all three methods, we use our IR and the tactile feedback for an appropriate comparison.

\begin{figure}[t]
    \centering
    \includegraphics[trim={0 0 0 0cm},clip,width=0.98\columnwidth]{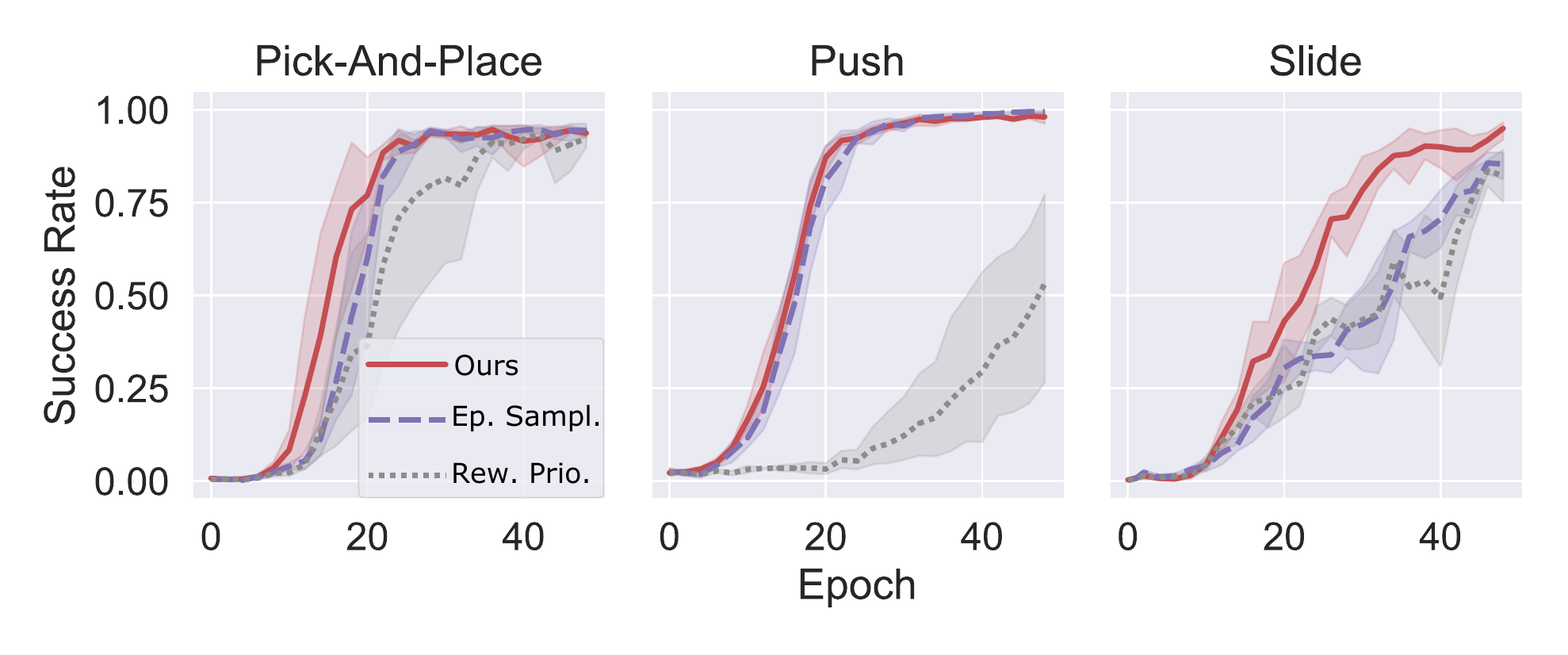}
    \vspace{-0.4cm}
    \caption{Ablation study of our sampling method. We show the average success rate over 5 random seeds with the corresponding confidence interval. 1 epoch corresponds to 40 training episodes. We compare our full method CPER + IR (red) and an ablation (purple) with a classical reward prioritiza-tion scheme (grey). In our ablation we find the training episodes according to CPER, but sample the state and virtual goal uniformly (like in HER).}
    \label{fig:sampling_ablation}
\end{figure}

Figure \ref{fig:sampling_ablation} illustrates the results. We can see that our full method is generally exploring faster. In particular, for \emph{Slide}, where the physical interactions with the object are limited to a few steps, we can see that \emph{episode sampling} shows slower convergence. This indicates that the full method is finding more informative transitions, which our method achieves by sampling virtual goals from $p_{\text{transition}}$ and then revert back to find a meaningful training transition. In this task, it is important to learn the swing trajectory prior to the kick and to choose virtual goals after the kick. If we choose transitions randomly, we more often sample non-informative transitions that occur after the kick or goals where the object does not move, which slows down training. The worse performance of the \emph{reward prioritization} baseline further indicates the benefit of our prioritization scheme.

\section{DISCUSSION}
\label{sec:discussion}

Here we address the current limitations of our method and propose areas for future work. This paper focuses on the three commonly used benchmarks in RL research for robotic manipulation tasks {\cite{plappert2018robotics}}. These fundamental problems have significantly different contact patterns and a single object in the scene. This allows us to demonstrate the benefits of an intrinsic reward based on tactile information and our proposed sampling scheme. However, applying our method to more complex tasks such as the manipulation of multiple objects or a door opening task, would require adding further modules. To solve such tasks, our method could be combined with other solutions, such as using a task-specific curriculum {\cite{curriculum2020}} for multiple objects or learning from expert demonstrations {\cite{rajeswaran2017learning}} for door opening.

To study the properties of the method, we conducted experiments in simulation. The deployment of simulator policies on a real robot often fails due to the discrepancy of sensor measurements and system dynamics between the real and simulated systems {\cite{hwangbo}}. To overcome this, either the accuracy of simulators needs to be improved or the policies have to be robust against imperfections in real world settings. Approaches that combine both improving the accuracy of simulators as well as applying domain randomization have been proposed {\cite{hwangbo}} and could be combined with our method. The contribution of our approach will most likely be robust for transfer to a real robot, since sensor noise will not trigger the threshold for receiving the intrinsic reward. One more aspect to consider is sensor placement, which we found crucial for avoiding exploitation of the intrinsic reward, e.g., by starting to push on the table. Here we use force sensors to get information about the contact with the object. In future work, one could infer tactile information from other sensor modalities, such as ultrasonic proximity sensors or vision \cite{vision2015}, which would further alleviate this issue and yield a simpler solution for a real system.

While we primarily focus on manipulation tasks, where using the sense of touch as intrinsic reward is intuitive, our method could be extended to other domains. In a more general sense, if a task can be decomposed into distinct phases, demarcated by a measurable event or physical property, the intrinsic reward and the sampling scheme could be applied. For example, reaching a certain amount of vertical thrust might be useful for learning flying skills in drones.

\section{CONCLUSION}

We study the challenges of exploration and efficient learning in deep reinforcement learning for robot manipulation tasks. We show in our experiments that an intrinsic force reward overcomes the difficulties of initial exploration and results in learning the intended behavior faster. We have discovered that transitions where the agent manipulates the object and brings it into motion contain valuable information for the learning process. We therefore introduce an up-sampling scheme that prioritizes exactly these transitions. We show that our prioritization scheme accelerates the learning progress even more. We find that our solution improves the performance and enhances the exploration on three fundamental manipulation tasks. Thus, we conclude that tactile feedback has the potential to advance reinforcement learning a step further.


\bibliographystyle{IEEEtran}
\bibliography{IEEEabrv, force}

\end{document}